\documentclass[runningheads]{llncs}
\usepackage[T1]{fontenc}
\usepackage{graphicx}
\usepackage{color}

\usepackage{algorithmicx}
\usepackage{algorithm, algpseudocode}
\usepackage{amsmath}
\usepackage{amsfonts}
\usepackage{amssymb}

\usepackage{makecell}
\usepackage{hyperref}
\usepackage{multirow}
\usepackage{mathrsfs}
\usepackage[symbol]{footmisc}

\begin{document}
\title{Pathology-and-genomics Multimodal Transformer for Survival Outcome Prediction}

\titlerunning{Pathology-and-genomics Multimodal Transformer for Survival Prediction}
%
\author{Kexin Ding\inst{1}\and Mu Zhou\inst{2}\and Dimitris N. Metaxas\inst{2}\and Shaoting Zhang\inst{3}}


\authorrunning{K. Ding et al.}

%

\institute{Department of Computer Science, UNC at Charlotte, North Carolina, USA \and Department of Computer Science, Rutgers University, New Jersey, USA \and Shanghai Artificial Intelligence Laboratory, Shanghai, China\\\email{zhangshaoting@pjlab.org.cn}}
%
%

%
\maketitle              

\begin{abstract}

Survival outcome assessment is challenging and inherently associated with multiple clinical factors (e.g., imaging and genomics biomarkers) in cancer. Enabling multimodal analytics promises to reveal novel predictive patterns of patient outcomes. In this study, we propose a multimodal transformer (\textbf{PathOmics}) integrating pathology and genomics insights into colon-related cancer survival prediction. We emphasize the unsupervised pretraining to capture the intrinsic interaction between tissue microenvironments in gigapixel whole slide images (WSIs) and a wide range of genomics data (e.g., mRNA-sequence, copy number variant, and methylation). After the multimodal knowledge aggregation in pretraining, our task-specific model finetuning could expand the scope of data utility applicable to both multi- and single-modal data (e.g., image- or genomics-only). We evaluate our approach on both TCGA colon and rectum cancer cohorts, showing that the proposed approach is competitive and outperforms state-of-the-art studies. Finally, our approach is desirable to utilize the limited number of finetuned samples towards data-efficient analytics for survival outcome prediction. The code is available at \url{https://github.com/Cassie07/PathOmics}.

\keywords{Histopathological image analysis \and Multimodal learning \and Cancer diagnosis \and Survival prediction}

\end{abstract}

\section{Introduction}

Cancers are a group of heterogeneous diseases reflecting deep interactions between pathological and genomics variants in tumor tissue environments~\cite{ref_article22}. Different cancer genotypes are translated into pathological phenotypes that could be assessed by pathologists~\cite{ref_article22}. High-resolution pathological images have proven their unique benefits for improving prognostic biomarkers prediction via exploring the tissue microenvironmental features~\cite{ref_article2,ref_article3,ref_article4,ref_article5,ref_article6,ref_article30}. Meanwhile, genomics data (e.g., mRNA-sequence) display a high relevance to regulate cancer progression~\cite{ref_article8,ref_article9}. For instance, genome-wide molecular portraits are crucial for cancer prognostic stratification and targeted therapy~\cite{ref_article27}. Despite their importance, seldom efforts jointly exploit the multimodal value between cancer image morphology and molecular biomarkers. In a broader context, assessing cancer prognosis is essentially a multimodal task in association with pathological and genomics findings. Therefore, synergizing multimodal data could deepen a cross-scale understanding towards improved patient prognostication.

The major goal of multimodal data learning is to extract complementary contextual information across modalities~\cite{ref_article1}. Supervised studies~\cite{ref_article12,ref_article13,ref_article11} have allowed multimodal data fusion among image and non-image biomarkers. For instance, the Kronecker product is able to capture the interactions between WSIs and genomic features for survival outcome prediction~\cite{ref_article12,ref_article13}. Alternatively, the co-attention transformer~\cite{ref_article11} could capture the genotype-phenotype interactions for prognostic prediction. Yet these supervised approaches are limited by feature generalizability and have a high dependency on data labeling. To alleviate label requirement, unsupervised learning evaluates the intrinsic similarity among multimodal representations for data fusion. For example, integrating image, genomics, and clinical information can be achieved via a predefined unsupervised similarity evaluation~\cite{ref_article1}. To broaden the data utility, the study ~\cite{ref_article10} leverages the pathology and genomic knowledge from the teacher model to guide the pathology-only student model for glioma grading. From these analyses, it is increasingly recognized that the lack of flexibility on model finetuning limits the data utility of multimodal learning. Meanwhile, the size of multimodal medical datasets is not as large as natural vision-language datasets, which necessitates the need for data-efficient analytics to address the training difficulty.

\begin{figure}[t!]
\includegraphics[width=\textwidth]{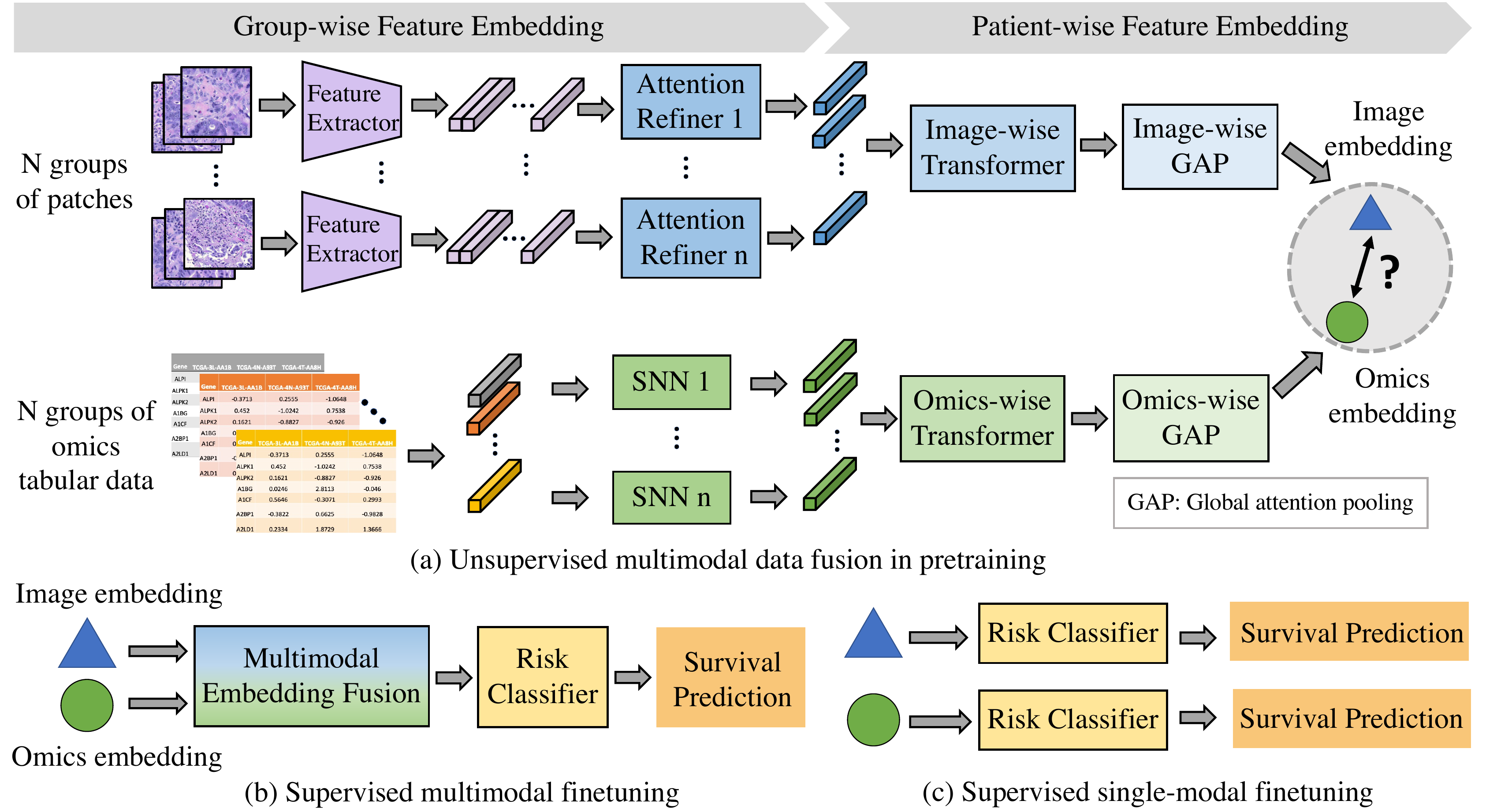}
\caption{Workflow overview of the pathology-and-genomics multimodal transformer (\textbf{PathOmics}) for survival prediction. In (a), we show the pipeline of extracting image and genomics feature embedding via an unsupervised pretraining towards multimodal data fusion. In (b) and (c), our supervised finetuning scheme could flexibly handle multiple types of data for prognostic prediction. With the multimodal pretrained model backbones, both multi- or single-modal data can be applicable for our model finetuning.}
\label{fig1}
\end{figure}

To tackle above challenges, we propose a pathology-and-genomics multimodal framework (i.e., \textbf{PathOmics}) for survival prediction (Fig~\ref{fig1}). We summarized our contributions as follows. \textbf{(1) Unsupervised multimodal data fusion.} Our unsupervised pretraining exploits the intrinsic interaction between morphological and molecular biomarkers (Fig~\ref{fig1}a). To overcome the gap of modality heterogeneity between images and genomics, we project the multimodal embeddings into the same latent space by evaluating the similarity among them. Particularly, the pretrained model offers a unique means by using similarity-guided modality fusion for extracting cross-modal patterns. \textbf{(2) Flexible modality finetuning.} A key contribution of our multimodal framework is that it combines benefits from both unsupervised pretraining and supervised finetuning data fusion (Fig~\ref{fig1}b). As a result, the task-specific finetuning broadens the dataset usage (Fig~\ref{fig1}b and c), which is not limited by data modality (e.g., both single- and multi-modal data). \textbf{(3) Data efficiency with limited data size.} Our approach could achieve comparable performance even with fewer finetuned data (e.g., only use 50\% of the finetuned data) when compared with using the entire finetuning dataset.

\section{Methodology}

\subsubsection{Overview.} Fig~\ref{fig1} illustrates our multimodal transformer framework. Our method includes an unsupervised multimodal data fusion pretraining and a supervised flexible-modal finetuning. From Fig~\ref{fig1}a, in the pretraining, our unsupervised data fusion aims to capture the interaction pattern of image and genomics features. Overall, we formulate the objective of multimodal feature learning by converting image patches and tabular genomics data into group-wise embeddings, and then extracting multimodal patient-wise embeddings. More specifically, we construct group-wise representations for both image and genomics modalities. For image feature representation, we randomly divide image patches into groups; Meanwhile, for each type of genomics data, we construct groups of genes depending on their clinical relevance~\cite{ref_article20}. Next, as seen in Fig~\ref{fig1}b and c, our approach enables three types of finetuning modal modes (i.e., multimodal, image-only, and genomics-only) towards prognostic prediction, expanding the downstream data utility from the pretrained model.

\subsubsection{Group-wise Image and Genomics Embedding.}
We define the group-wise genomics representation by referring to $N=8$ major functional groups obtained from~\cite{ref_article20}. Each group contains a list of well-defined molecular features related to cancer biology, including transcription factors, tumor suppression, cytokines and grow\-th factors, cell differentiation markers, homeodomain proteins, translocated cancer genes, and protein kinases. The group-wise genomics representation is defined as $G_{n} \in \mathbb{R}^{1\times d_{g}}$, where $n \in N$, $d_{g}$ is the attribute dimension in each group which could be various. To better extract high-dimensional group-wise genomics representation, we use a Self-Normalizing Network (SNN) together with scaled exponential linear units (SeLU) and Alpha Dropout for feature extraction to generate the group-wise embedding $G_{n} \in \mathbb{R}^{1\times 256}$ for each group.

For group-wise WSIs representation, we first cropped all tissue-region image tiles from the entire WSI and extracted CNN-based (e.g., ResNet50) $d_{i}$-dimensional features for each image tile k as $h_{k} \in \mathbb{R}^{1 \times d_{i}}$, where $d_{i}=1,024$, $k \in K$ and K is the number of image patches. We construct the group-wise WSIs representation by randomly splitting image tile features into N groups (i.e., the same number as genomics categories). Therefore, group-wise image representation could be defined as $I_{n} \in \mathbb{R}^{k_{n}\times1024}$, where $n \in N$ and $k_{n}$ represents tile k in group n. Then we apply an attention-based refiner (ABR)~\cite{ref_article21}, which is able to weight the feature embeddings in the group, together with a dimension deduction (e.g., fully-connected layers) to achieve the group-wise embedding. The ABR and the group-wise embedding $I_{n} \in \mathbb{R}^{1\times256}$ are defined as:

\begin{equation}
a_{k} = \frac{epx\{ w^{T}(tanh(V_{1}h_{k})\odot (sigm(V_{2}h_{k}))\}}{\sum_{j=1}^{K} epx\{ w^{T}(tanh(V_{1}h_{j})\odot (sigm(V_{2}h_{j}))\}}
\label{eq1}
\end{equation}
where w,V1 and V2 are the learnable parameters.

\begin{equation}
I_{n} = \sum_{k=1}^{K} a_{k}h_{k}
\label{eq2}
\end{equation}

\subsubsection{Patient-wise Multimodal Feature Embedding.}
To aggregate patient-wise multimodal feature embedding from the group-wise representations, as shown in Fig~\ref{fig1}a, we propose a pathology-and-genomics multimodal model containing two model streams, including a pathological image and a genomics data stream. In each stream, we use the same architecture with different weights, which is updated separately in each modality stream. In the pathological image stream, the patient-wise image representation is aggregated by N group representations as $I_{p} \in \mathbb{R}^{N\times 256}$, where $p \in P$ and P is the number of patients. Similarly, the patient-wise genomics representation is aggregated as $G_{p} \in \mathbb{R}^{N\times 256}$. After generating patient-wise representation, we utilize two transformer layers~\cite{ref_article24} to extract feature embeddings for each modality as follows:

\begin{equation}
H_{p}^{l} = MSA(H_{p})
\end{equation}

where MSA denotes Multi-head Self-attention~\cite{ref_article24} (see Appendix 1), l denotes the layer index of the transformer, and $H_{p}$ could either be $I_{p}$ or $G_{p}$. Then, we construct global attention poolings~\cite{ref_article21} as Eq.\ref{eq1} to adaptively compute a weighted sum of each modality feature embeddings to finally construct patient-wise embedding as $I_{embedding}^{p} \in \mathbb{R}^{1\times256}$ and $G_{embedding}^{p} \in \mathbb{R}^{1\times256}$ in each modality. 

\subsubsection{Multimodal Fusion in Pretraining and Finetuning.} Due to the domain gap between image and molecular feature heterogeneity, a proper design of multimodal fusion is crucial to advance integrative analysis. In the pretraining stage, we develop an unsupervised data fusion strategy by decreasing the mean square error (MSE) loss to map images and genomics embeddings into the same space. Ideally, the image and genomics embeddings belonging to the same patient should have a higher relevance between each other. MSE measures the average squared difference between multimodal embeddings. In this way, the pretrained model is trained to map the paired image and genomics embeddings to be closer in the latent space, leading to strengthen the interaction between different modalities.

\begin{equation}
\mathcal{L}_{fusion} = argmin \frac{1}{P}\sum_{p=1}^{P}((I_{embedding}^{p} - G_{embedding}^{p})^{2})
\end{equation}
In the single modality finetuning, even if we use image-only data, the model is able to produce genomic-related image feature embedding due to the multimodal knowledge aggregation already obtained from the model pretraining. As a result, our cross-modal information aggregation relaxes the modality requirement in the finetuning stage. As shown in Fig~\ref{fig1}b, for multimodal finetuning, we deploy a concatenation layer to obtain the fused multimodal feature representation and implement a risk classifier (FC layer) to achieve the final survival stratification (see Appendix 2). As for single-modality finetuning mode in Fig~\ref{fig1}c, we simply feed $I_{embedding}^{p}$ or $G_{embedding}^{p}$ into risk classifier for the final prognosis prediction. During the finetuning, we update the model parameters using a log-likelihood loss for the discrete-time survival model training~\cite{ref_article11}(see Appendix 2).

\section{Experiments and Results}
\subsubsection{Datasets.} 

All image and genomics data are publicly available. We collected WSIs from The Cancer Genome Atlas Colon Adenocarcinoma (TCGA-COAD) dataset (CC-BY-3.0)~\cite{ref_article14,ref_article29} and Rectum Adenocarcinoma (TCGA-READ) dataset (CC-BY-3.0)~\cite{ref_article16,ref_article29}, which contain 440 and 153 patients. We cropped each WSI into 512 $\times$ 512 non-overlapped patches. We also collected the corresponding tabular genomics data (e.g., mRNA sequence, copy number alteration, and methylation) with overall survival (OS) times and censorship statuses from Cbioportal~\cite{ref_article17,ref_article18}. We removed the samples without the corresponding genomics data or ground truth of survival outcomes. Finally, we included 426 patients of TCGA-COAD and 145 patients of TCGA-READ.

\begin{figure}[t!]
\includegraphics[width=\textwidth]{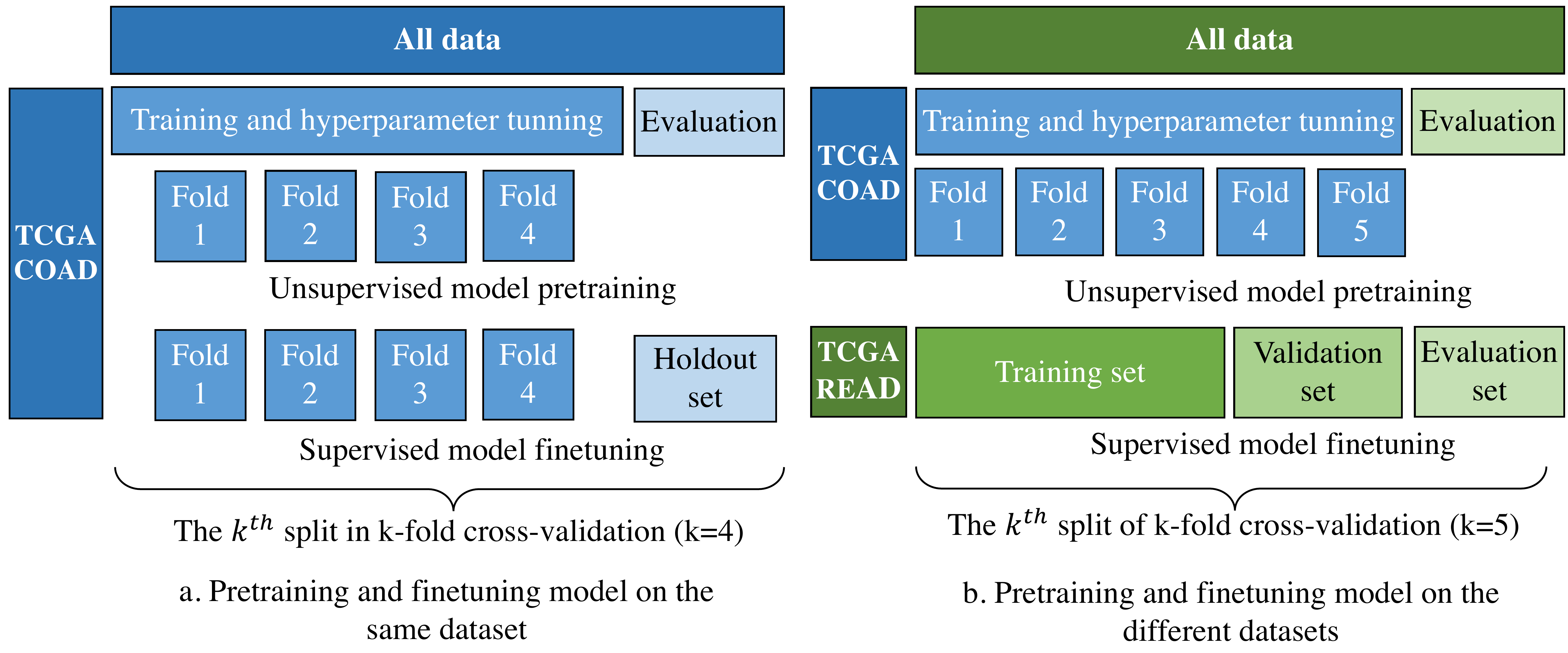}
\caption{Dataset usage. In a, we use TCGA-COAD dataset for model pretraining, finetuning, and evaluation. In b, we use TCGA-COAD dataset for model pretraining. Then, we use TCGA-READ dataset to finetune and evaluate the pretrained models.}
\label{fig2}
\end{figure}

\subsubsection{Experimental Settings and Implementations.} 
We implement two types of settings that involve internal and external datasets for model pretraining and finetuning. As shown in Fig~\ref{fig2}a, we pretrain and finetune the model on the same dataset (i.e., internal setting). We split TCGA-COAD into training (80\%) and holdout testing set (20\%). Then, we implement four-fold cross-validation on the training set for pretraining, finetuning, and hyperparameter-tuning. The test set is only used for evaluating the best finetuned models from each cross-validation split. For the external setting, we implement pretraining and finetuning on the different datasets, as shown in Fig~\ref{fig2}b; we use TCGA-COAD for pretraining; Then, we only use TCGA-READ for finetuning and final evaluation. We implement a five-fold cross-validation for pretraining, and the best pretrained models are used for finetuning. We split TCGA-READ into finetuning (60\%), validation (20\%), and evaluation set (20\%). For all experiments, we calculate the average performance on the evaluation set across the best models. 

The number of epochs for pretraining and finetuning are 25, the batch size is 1, the optimizer is Adam~\cite{ref_article19}, and the learning rate is 1e-4 for pretraining and 5e-5 for finetuning. We used one 32GB Tesla V100 SXM2 GPU and Pytorch. The concordance index (C-index) is used to measure the survival prediction performance. We followed the previous studies~\cite{ref_article11,ref_article12,ref_article13} to partition the overall survival (OS) months into four non-overlapping intervals by using the quartiles of event times of uncensored patients for discretized-survival C-index calculation (see Appendix 2). For each experiment, we reported the average C-index among three-times repeated experiments. Conceptionally, our method shares a similar idea to multiple instance learning (MIL)~\cite{ref_article25,ref_article26}. Therefore, we include two types of baseline models, including the MIL-based models (DeepSet~\cite{ref_article23}, AB-MIL~\cite{ref_article21}, and TransMIL~\cite{ref_article28}) and MIL multimodal-based models (MCAT~\cite{ref_article11}, PORPOISE~\cite{ref_article13}). We follow the same data split and processing, as well as the identical training hyperparameters and supervised fusion as above. Notably, there is no need for supervised finetuning for the baselines when using TCGA-COAD (Table~\ref{tab1}), because the supervised pretraining is already applied to the training set.

\begin{table}[t!]
\centering
\caption{The comparison of C-index performance on TCGA-COAD and TCGA-READ dataset. "Methy" is used as the abbreviation of Methylation.}\label{tab1}
\begin{tabular}{c|c|c|c|c|c}
\hline

Model & \makecell{Pretrain\\data modality} & \multicolumn{2}{c|}{TCGA-COAD} & \multicolumn{2}{c}{TCGA-READ}\\
 \cline{3-6}
 &  & \makecell{Finetune\\data modality} & C-index (\%) & \makecell{Finetune\\data modality} & C-index (\%)\\
\hline
 & image+mRNA & - & $58.70\pm 1.10 $ & image+mRNA &$70.19\pm1.45$\\
\makecell{DeepSets\\~\cite{ref_article23}} & image+CNA & - & $51.50\pm 2.60$& image+CNA & $62.50\pm2.52$\\
 & image+Methy & - & $65.61\pm1.86$& \makecell{image+Methy} & $55.78\pm1.22$\\
\hline
 & image+mRNA &  - & $54.12\pm 2.88$& image+mRNA & $68.79\pm1.44$\\
\makecell{AB-MIL\\~\cite{ref_article21}} & image+CNA & - & $54.68\pm 2.44$ & image+CNA & $66.72\pm0.81$\\
 & image+Methy & - & $49.66\pm 1.58$& image+Methy & $55.78\pm1.22$\\
\hline
 & image+mRNA & -  & $54.15 \pm1.02$ & image+mRNA  & $67.91\pm2.35$\\
\makecell{TransMIL\\~\cite{ref_article28}} & image+CNA& -  & $59.80\pm0.98$ & image+CNA & $62.75\pm1.92$\\
 & image+Methy & - & $53.35\pm1.78$ & image+Methy & $53.09\pm1.46$\\
\hline
 & image+mRNA &  - & $65.02\pm 3.10$& image+mRNA & $70.27\pm2.75$\\
\makecell{MCAT\\~\cite{ref_article11}} & image+CNA & - &$64.66\pm 2.31$& image+CNA &$60.50\pm1.25$\\
 & image+Methy & - & $60.98\pm2.43$& image+Methy & $59.78\pm1.20$\\
\hline
 & image+mRNA & -  & $65.31\pm1.26$ & image+mRNA  & $68.18\pm1.62$\\
\makecell{PORPOI\\-SE~\cite{ref_article13}} & image+CNA& -  & $57.32\pm1.78$ & image+CNA & $60.19\pm1.48$\\
 & image+Methy & - & $61.84\pm1.10$ & image+Methy & $68.80\pm0.92$\\
\hline
\hline
 \multirow{9}{*}{Ours} & \multirow{3}{*}{image+mRNA}  & image+mRNA & $\mathbf{67.32\pm 1.69}$ & image+mRNA& $74.35\pm1.15$\\
 &  & image    & $63.78\pm1.22$& image& $\mathbf{74.85\pm 0.37}$\\
 &  & mRNA    & $60.76\pm0.88$& mRNA & $59.61\pm1.37$\\
\cline{2-6}
 &\multirow{3}{*}{image+CNA}& image+CNA & $61.19\pm 1.03$ & image+CNA & $73.95\pm1.05$\\
 &  & image    & $58.06\pm1.54$& image & $71.18\pm1.39$\\
 &  & CNA    & $56.43\pm1.02$& CNA & $63.95\pm0.55$\\
\cline{2-6}
 & \multirow{3}{*}{image+Methy}& image+Methy & $67.22\pm1.67$& image+Methy & $71.80\pm2.03$\\
 &  & image & $60.43\pm0.72$& image & $64.42\pm0.72$\\
 &  & Methy    & $61.06\pm1.34$ & Methy & $65.42\pm0.91$\\
\hline

\end{tabular}
\end{table} 

\subsubsection{Results.}
In Table~\ref{tab1}, our approach shows improved survival prediction performance on both TCGA-COAD and TCGA-READ datasets. Compared with supervised baselines, our unsupervised data fusion is able to extract the phenotype-genotype interaction features, leading to achieving a flexible finetuning for different data settings. With the multimodal pretraining and finetuning, our method outperforms state-of-the-art models by about 2\% on TCGA-COAD and 4\% TCGA-READ. We recognize that the combination of image and mRNA sequencing data leads to reflecting distinguishing survival outcomes. Remarkably, our model achieved positive results even using a single-modal finetuning when compared with baselines (more results in Appendix 3.1). In the meantime, on the TCGA-READ, our single-modality finetuned model achieves a better performance than multimodal finetuned baseline models (e.g., with model pretraining via image and methylation data, we have only used the image data for finetuning and achieved a C-index of 74.85\%, which is about 4\% higher than the best baseline models). We show that with a single-modal finetuning strategy, the model could generate meaningful embedding to combine image- and genomic-related patterns. In addition, our model reflects its efficiency on the limited finetuning data (e.g., 75 patients are used for finetuning on TCGA-READ, which are only 22\% of TCGA-COAD finetuning data). In Table~\ref{tab1}, our method could yield better performance compared with baselines on the small dataset across the combination of images and multiple types of genomics data.

\begin{figure}[t!]
\includegraphics[width=\textwidth]{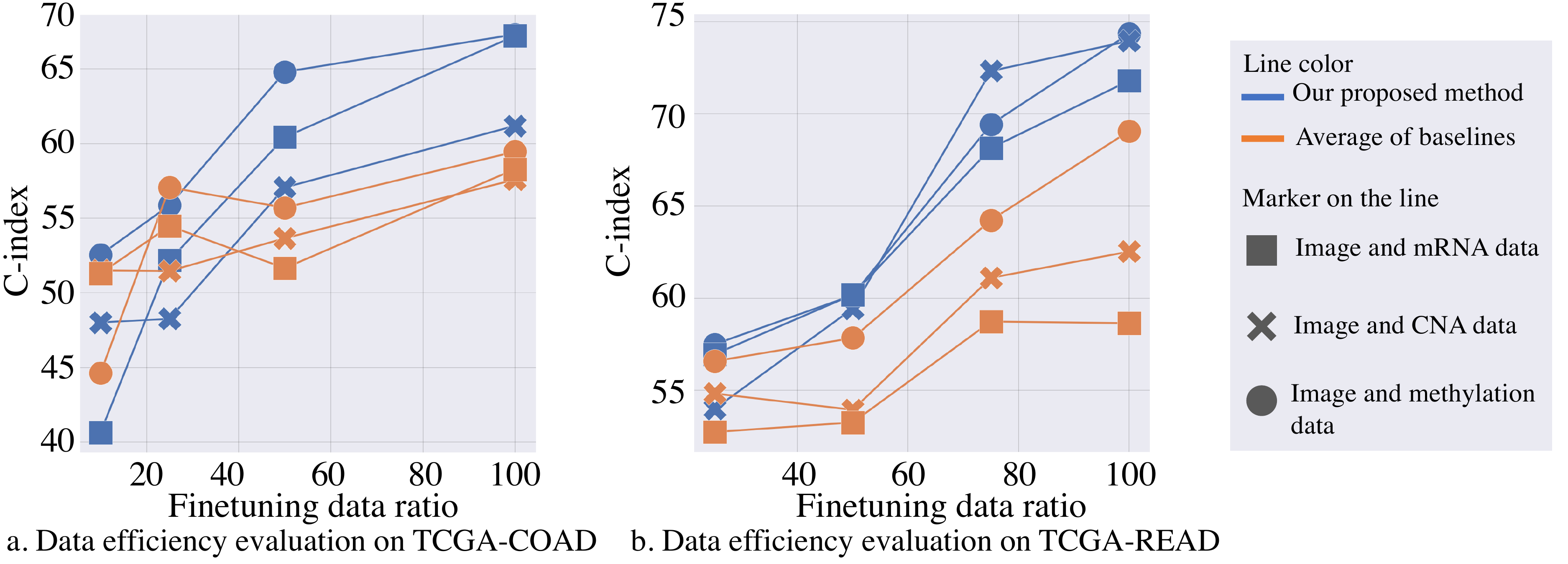}
\caption{Ablation study. In (a) and (b), we evaluate the model efficiency by using fewer data for model finetuning on TCGA-COAD and TCGA-READ. We show the average C-index of baselines, the detailed results are shown in the Appendix 3.2.}
\label{fig3}
\end{figure}

\subsubsection{Ablation Analysis.}
We verify the model efficiency by using fewer amounts of finetuning data in finetuning. For TCGA-COAD dataset, we include 50\%, 25\%, and 10\% of the finetuning data. For the TCGA-READ dataset, as the number of uncensored patients is limited, we use 75\%, 50\%, and 25\% of the finetuning data to allow at least one uncensored patient to be included for finetuning. As shown in Fig~\ref{fig3}a, by using 50\% of TCGA-COAD finetuning data, our approach achieves the C-index of 64.80\%, which is higher than the average performance of baselines in several modalities. Similarly, in Fig~\ref{fig3}b, our model retains a good performance by using 50\% or 75\% of TCGA-READ finetuning data compared with the average of C-index across baselines (e.g., 72.32\% versus 64.23\%). For evaluating the effect of cross-modality information extraction in the pretraining, we kept supervised model training (i.e., the finetuning stage) while removing the unsupervised pretraining. The performance is lower 2\%-10\% than ours on multi- and single-modality data. For evaluating the genomics data usage, we designed two settings: (1) combining all types of genomics data and categorizing them by groups; (2) removing category information while keeping using different types of genomics data separately. Our approach outperforms the above ablation studies by 3\%-7\% on TCGA-READ and performs similarly on TCGA-COAD. In addition, we replaced our unsupervised loss with cosine similarity loss; our approach outperforms the setting of using cosine similarity loss by 3\%-6\%.

\section{Conclusion}
Developing data-efficient multimodal learning is crucial to advance the survival assessment of cancer patients in a variety of clinical data scenarios. We demonstrated that the proposed PathOmics framework is useful for improving the survival prediction of colon and rectum cancer patients. Importantly, our approach opens up perspectives for exploring the key insights of intrinsic genotype-phenotype interactions in complex cancer data across modalities. Our finetuning approach broadens the scope of dataset inclusion, particularly for model finetuning and evaluation, while enhancing model efficiency on analyzing multimodal clinical data in real-world settings. In addition, the use of synthetic data and developing a foundation model training will be helpful to improve the robustness of multimodal data fusion~\cite{ref_article31,ref_article32}.


\subsubsection{Acknowledgements.} The results of this study are based on the data collected from the public TCGA Research Network: \url{https://www.cancer.gov/tcga}.

%
%
\bibliographystyle{splncs04}
\bibliography{Reference1847}

\begin{thebibliography}{10}
\providecommand{\url}[1]{\texttt{#1}}
\providecommand{\urlprefix}{URL }
\providecommand{\doi}[1]{https://doi.org/#1}

\bibitem{ref_article4}
Bilal, M., Raza, S.E.A., Azam, A., Graham, S., Ilyas, M., Cree, I.A., Snead,
  D., Minhas, F., Rajpoot, N.M.: Development and validation of a weakly
  supervised deep learning framework to predict the status of molecular
  pathways and key mutations in colorectal cancer from routine histology
  images: a retrospective study. The Lancet Digital Health  \textbf{3}(12),
  e763--e772 (2021)

\bibitem{ref_article17}
Cerami, E., Gao, J., Dogrusoz, U., Gross, B.E., Sumer, S.O., Aksoy, B.A.,
  Jacobsen, A., Byrne, C.J., Heuer, M.L., Larsson, E., et~al.: The cbio cancer
  genomics portal: an open platform for exploring multidimensional cancer
  genomics data. Cancer discovery  \textbf{2}(5),  401--404 (2012)

\bibitem{ref_article9}
Chaudhary, K., Poirion, O.B., Lu, L., Garmire, L.X.: Deep learning--based
  multi-omics integration robustly predicts survival in liver cancerusing deep
  learning to predict liver cancer prognosis. Clinical Cancer Research
  \textbf{24}(6),  1248--1259 (2018)

\bibitem{ref_article1}
Cheerla, A., Gevaert, O.: Deep learning with multimodal representation for
  pancancer prognosis prediction. Bioinformatics  \textbf{35}(14),  i446--i454
  (2019)

\bibitem{ref_article12}
Chen, R.J., Lu, M.Y., Wang, J., Williamson, D.F., Rodig, S.J., Lindeman, N.I.,
  Mahmood, F.: Pathomic fusion: an integrated framework for fusing
  histopathology and genomic features for cancer diagnosis and prognosis. IEEE
  Transactions on Medical Imaging  \textbf{41}(4),  757--770 (2020)

\bibitem{ref_article11}
Chen, R.J., Lu, M.Y., Weng, W.H., Chen, T.Y., Williamson, D.F., Manz, T.,
  Shady, M., Mahmood, F.: Multimodal co-attention transformer for survival
  prediction in gigapixel whole slide images. In: Proceedings of the IEEE/CVF
  International Conference on Computer Vision. pp. 4015--4025 (2021)

\bibitem{ref_article13}
Chen, R.J., Lu, M.Y., Williamson, D.F., Chen, T.Y., Lipkova, J., Noor, Z.,
  Shaban, M., Shady, M., Williams, M., Joo, B., et~al.: Pan-cancer integrative
  histology-genomic analysis via multimodal deep learning. Cancer Cell
  \textbf{40}(8),  865--878 (2022)

\bibitem{ref_article29}
Clark, K., Vendt, B., Smith, K., Freymann, J., Kirby, J., Koppel, P., Moore,
  S., Phillips, S., Maffitt, D., Pringle, M., et~al.: The cancer imaging
  archive (tcia): maintaining and operating a public information repository.
  Journal of digital imaging  \textbf{26},  1045--1057 (2013)

\bibitem{ref_article25}
Dietterich, T.G., Lathrop, R.H., Lozano-P{\'e}rez, T.: Solving the multiple
  instance problem with axis-parallel rectangles. Artificial intelligence
  \textbf{89}(1-2),  31--71 (1997)

\bibitem{ref_article3}
Ding, K., Liu, Q., Lee, E., Zhou, M., Lu, A., Zhang, S.: Feature-enhanced graph
  networks for genetic mutational prediction using histopathological images in
  colon cancer. In: Medical Image Computing and Computer Assisted
  Intervention--MICCAI 2020: 23rd International Conference, Lima, Peru, October
  4--8, 2020, Proceedings, Part II 23. pp. 294--304. Springer (2020)

\bibitem{ref_article31}
Ding, K., Zhou, M., Wang, H., Gevaert, O., Metaxas, D., Zhang, S.: A
  large-scale synthetic pathological dataset for deep learning-enabled
  segmentation of breast cancer. Scientific Data  \textbf{10}(1), ~231 (2023)

\bibitem{ref_article5}
Ding, K., Zhou, M., Wang, H., Zhang, S., Metaxas, D.N.: Spatially aware graph
  neural networks and cross-level molecular profile prediction in colon cancer
  histopathology: a retrospective multi-cohort study. The Lancet Digital Health
   \textbf{4}(11),  e787--e795 (2022)

\bibitem{ref_article30}
Ding, K., Zhou, M., Wang, Z., Liu, Q., Arnold, C.W., Zhang, S., Metaxas, D.N.:
  Graph convolutional networks for multi-modality medical imaging: Methods,
  architectures, and clinical applications. arXiv preprint arXiv:2202.08916
  (2022)

\bibitem{ref_article18}
Gao, J., Aksoy, B.A., Dogrusoz, U., Dresdner, G., Gross, B., Sumer, S.O., Sun,
  Y., Jacobsen, A., Sinha, R., Larsson, E., et~al.: Integrative analysis of
  complex cancer genomics and clinical profiles using the cbioportal. Science
  signaling  \textbf{6}(269),  pl1--pl1 (2013)

\bibitem{ref_article32}
Gao, Y., Li, Z., Liu, D., Zhou, M., Zhang, S., Meta, D.N.: Training like a
  medical resident: Universal medical image segmentation via context prior
  learning. arXiv preprint arXiv:2306.02416  (2023)

\bibitem{ref_article27}
Gentles, A.J., Newman, A.M., Liu, C.L., Bratman, S.V., Feng, W., Kim, D., Nair,
  V.S., Xu, Y., Khuong, A., Hoang, C.D., et~al.: The prognostic landscape of
  genes and infiltrating immune cells across human cancers. Nature medicine
  \textbf{21}(8),  938--945 (2015)

\bibitem{ref_article21}
Ilse, M., Tomczak, J., Welling, M.: Attention-based deep multiple instance
  learning. In: International conference on machine learning. pp. 2127--2136.
  PMLR (2018)

\bibitem{ref_article2}
Kather, J.N., Pearson, A.T., Halama, N., J{\"a}ger, D., Krause, J., Loosen,
  S.H., Marx, A., Boor, P., Tacke, F., Neumann, U.P., et~al.: Deep learning can
  predict microsatellite instability directly from histology in
  gastrointestinal cancer. Nature medicine  \textbf{25}(7),  1054--1056 (2019)

\bibitem{ref_article19}
Kingma, D.P., Ba, J.: Adam: A method for stochastic optimization. arXiv
  preprint arXiv:1412.6980  (2014)

\bibitem{ref_article16}
Kirk, S., Lee, Y., Sadow, C., Levine: The cancer genome atlas rectum
  adenocarcinoma collection (tcga-read) (version 3) [data set]. The Cancer
  Imaging Archive  (2016)

\bibitem{ref_article14}
Kirk, S., Lee, Y., Sadow, C., Levine, S., Roche, C., Bonaccio, E., Filiippini,
  J.: Radiology data from the cancer genome atlas colon adenocarcinoma
  [tcga-coad] collection. The Cancer Imaging Archive  (2016)

\bibitem{ref_article20}
Liberzon, A., Birger, C., Thorvaldsd{\'o}ttir, H., Ghandi, M., Mesirov, J.P.,
  Tamayo, P.: The molecular signatures database hallmark gene set collection.
  Cell systems  \textbf{1}(6),  417--425 (2015)

\bibitem{ref_article26}
Maron, O., Lozano-P{\'e}rez, T.: A framework for multiple-instance learning.
  Advances in neural information processing systems  \textbf{10} (1997)

\bibitem{ref_article22}
Marusyk, A., Almendro, V., Polyak, K.: Intra-tumour heterogeneity: a looking
  glass for cancer? Nature reviews cancer  \textbf{12}(5),  323--334 (2012)

\bibitem{ref_article6}
Qu, H., Zhou, M., Yan, Z., Wang, H., Rustgi, V.K., Zhang, S., Gevaert, O.,
  Metaxas, D.N.: Genetic mutation and biological pathway prediction based on
  whole slide images in breast carcinoma using deep learning. NPJ precision
  oncology  \textbf{5}(1), ~87 (2021)

\bibitem{ref_article28}
Shao, Z., Bian, H., Chen, Y., Wang, Y., Zhang, J., Ji, X., et~al.: Transmil:
  Transformer based correlated multiple instance learning for whole slide image
  classification. Advances in neural information processing systems
  \textbf{34},  2136--2147 (2021)

\bibitem{ref_article24}
Vaswani, A., Shazeer, N., Parmar, N., Uszkoreit, J., Jones, L., Gomez, A.N.,
  Kaiser, {\L}., Polosukhin, I.: Attention is all you need. Advances in neural
  information processing systems  \textbf{30} (2017)

\bibitem{ref_article10}
Xing, X., Chen, Z., Zhu, M., Hou, Y., Gao, Z., Yuan, Y.: Discrepancy and
  gradient-guided multi-modal knowledge distillation for pathological glioma
  grading. In: Medical Image Computing and Computer Assisted
  Intervention--MICCAI 2022: 25th International Conference, Singapore,
  September 18--22, 2022, Proceedings, Part V. pp. 636--646. Springer (2022)

\bibitem{ref_article8}
Yang, M., Yang, H., Ji, L., Hu, X., Tian, G., Wang, B., Yang, J.: A multi-omics
  machine learning framework in predicting the survival of colorectal cancer
  patients. Computers in Biology and Medicine  \textbf{146},  105516 (2022)

\bibitem{ref_article23}
Zaheer, M., Kottur, S., Ravanbakhsh, S., Poczos, B., Salakhutdinov, R.R.,
  Smola, A.J.: Deep sets. Advances in neural information processing systems
  \textbf{30} (2017)

\end{thebibliography}

%



%
%
%
\title{Appendix}

%
%

\author{Kexin Ding\inst{1}\and Mu Zhou\inst{2}\and Dimitris N. Metaxas\inst{2}\and Shaoting Zhang\inst{3}}


\authorrunning{K. Ding et al.}

%

\institute{Department of Computer Science, UNC at Charlotte, North Carolina, USA \and Department of Computer Science, Rutgers University, New Jersey, USA \and Shanghai Artificial Intelligence Laboratory, Shanghai, China\\}



%

%
%
%
%
\maketitle              
\renewcommand{\thetable}{S\arabic{table}}
\section{Multi-head Self-attention}
The Multihead self-attention (MSA) is the concatenation of k self-attention (SA) operations. The SA uses $d_{k}$-dim patient embedding as the query Q, key K, and value V to learn the pairwise relationship $a_{ij} \in A$ between $q_{i} \in Q $ and $k_{i} \in K $:

\begin{equation}
softmax(\frac{QK^{T}}{\sqrt{d_{k}}}) = A
\label{eqS1}
\end{equation}

\begin{equation}
SA(Q,K,V) = AV
\label{eqS2}
\end{equation}

\section{Discrete-time Survival Prediction and Evaluation}
We extend the definition and detailed proof of "discrete-time survival prediction" as follows. The continuous event time $T_{j,continue} \in [t_r, t_r+1)$ could be discretized as $T_{j}$, which is equal to r, where $r \in \{0, 1, 2, 3\}$ and j is the index of four non-overlapped intervals. The discrete ground truth is $Y_{j}\in \{0, 1, 2, 3\}$. With patient-wise embedding $h_{final_j}$, we define the hazard function $f_{hazard}(r| h_{final_j})$ as $P(T_{j}=r|T_{j}\geq r, h_{final_j})$, which is used for calculating the survival function (i.e., C-index calculation) $f_{surv}(r| h_{final_j})$ through $P(T_{j}>r|h_{final_j})$ (i.e., $\prod_{u=1}^{r}(1-f_{hazard}(u|h_{final_j}))$). During the supervised finetuning, the log-likelihood loss for model parameter updation is defined as $-c_{j}\cdot log(f_{surv}(Y_{j}|h_{final_j}))-(1-c_{j})\cdot log(f_{surv}(Y_{j}-1|h_{final_j}))-(1-c_{j})\cdot log(f_{hazard}(Y_{j}|h_{final_j}))$, where $c_{j} = 0$ means patient passed away during $T_{j}$ and $c_{j} = 1$ means patient lived after $T_{j}$.

\section{Appendix Results of Baseline Models}
\subsection{The Results of Baseline Models with Single-modal Data}
In Table~\ref{tabS1}, We reported the results of the baselines with single modality data. Each column represents the C-index values on the testing set in single modality.

\begin{table}[t!]
\centering
\caption{Single-modal finetuning and testing on TCGA-COAD and TCGA-READ.}\label{tabS1}
\begin{tabular}{c|c|c|c|c|c|c}
\hline
\multicolumn{7}{c}{TCGA-COAD}\\
\hline
Model & \multicolumn{2}{c|}{\makecell{Pretrain with \\image and mRNA}} & \multicolumn{2}{c|}{\makecell{Pretrain with \\image and CNA}} & \multicolumn{2}{c}{\makecell{Pretrain with \\image and methylation}}\\
 \cline{2-7}
 & Image & mRNA & Image & CNA & Image & Methy\\
 \hline
\makecell{Deep\\-Sets} & $52.95\pm2.27 $ & $55.17\pm3.19 $& $50.94\pm2.28 $& $58.78\pm2.11 $& $55.71\pm2.02 $& $54.18\pm2.86 $\\
\hline
\makecell{AB\\-MIL} & $56.82\pm2.73 $ & $60.63\pm2.56 $& $54.68\pm2.38 $& $54.18\pm2.51 $& $55.02\pm2.32 $& $55.78\pm2.43$\\
\hline
\makecell{Trans\\-MIL} & $59.68\pm1.65 $ & $58.72\pm1.21 $& $61.28\pm1.66 $& $50.27\pm2.27 $& $55.83\pm1.75 $& $50.73\pm2.02 $\\
\hline
MCAT & $60.32\pm2.57 $ & $58.63\pm2.29 $& $59.24\pm1.93 $& $55.01\pm2.74 $& $60.85\pm1.74 $& $61.11\pm2.25 $\\
\hline
\makecell{PORP\\-OISE} & $57.50\pm1.83 $ & $60.44\pm2.71 $& $55.87\pm2.88 $& $51.76\pm2.05 $& $43.58\pm2.15 $& $55.92\pm2.65 $\\
\hline
\multicolumn{7}{c}{TCGA-READ}\\
\hline
\makecell{Deep\\-Sets} & $60.45\pm2.49 $ & $61.33\pm1.64 $& $56.59\pm2.72 $& $50.47\pm2.19 $& $57.82\pm2.14 $& $60.01\pm2.23 $\\
\hline
\makecell{AB\\-MIL} & $68.73\pm1.98 $ & $56.58\pm1.86 $& $62.63\pm1.62 $& $60.57\pm2.24 $& $61.63\pm2.19$& $53.49\pm2.31 $\\
\hline
\makecell{Trans\\-MIL} & $50.59\pm2.49 $ & $50.33\pm2.67 $& $60.92\pm2.82 $& $55.46\pm2.19 $& $51.48\pm1.52 $& $60.47\pm1.82 $\\
\hline
MCAT & $66.95\pm2.51 $ & $62.41\pm2.05 $& $58.73\pm2.38 $& $63.77\pm2.03 $& $62.78\pm2.15 $& $58.87\pm2.09 $\\
\hline
\makecell{PORP\\-OISE} & $62.42\pm1.84$ & $58.49\pm1.60 $& $56.87\pm2.18 $& $58.41\pm2.20 $& $67.60\pm2.26 $& $66.38\pm2.74 $\\
\hline
\end{tabular}
\end{table}

\begin{table}[t!]
\centering
\caption{Fewer data finetuning performance on TCGA-COAD and TCGA-READ.}\label{tabS2}
\begin{tabular}{c|c|c|c|c|c|c}
\hline

\makecell{Model} & \multicolumn{3}{c|}{TCGA-COAD} & \multicolumn{3}{c}{TCGA-READ}\\
 \cline{2-7}
 & 50\% & 25\% & 10\% & 75\% & 50\% & 25\%\\
\hline
 & $63.16\pm 2.23 $ & $62.76\pm2.54 $& $40.39\pm3.11 $& $69.05\pm2.31 $& $55.55\pm2.05 $& $47.12\pm2.82 $\\
\makecell{Deep\\-Sets} & $58.91\pm2.44 $ & $58.31\pm2.01 $& $49.50\pm2.25 $& $60.90\pm2.34 $& $55.45\pm2.51 $& $49.77\pm2.19$\\
 & $45.37\pm3.42 $ & $54.88\pm3.27 $& $54.23\pm2.76 $& $64.26\pm1.98 $& $51.38\pm2.23 $& $49.65\pm2.62 $\\
\hline
 & $51.35\pm2.31$ & $54.64\pm1.82 $& $45.26\pm1.91 $& $66.66\pm2.43 $& $48.78\pm1.82 $& $49.35\pm2.61 $\\
\makecell{AB\\-MIL} & $53.02\pm2.44 $ & $44.93\pm2.84 $& $47.40\pm2.23 $& $62.93\pm2.30 $& $62.90\pm2.04 $& $62.37\pm2.54 $\\
 & $52.20\pm2.00 $ & $54.50\pm2.91 $& $50.34\pm2.55 $& $51.87\pm2.66 $& $50.71\pm2.31 $& $50.33\pm2.86 $\\
\hline
 & $53.25\pm1.92$ & $59.06\pm2.53 $& $45.10\pm2.40 $& $59.32\pm2.92 $& $63.25\pm2.57 $& $63.63\pm2.82 $\\
\makecell{Trans\\-MIL} & $48.62\pm2.61 $ & $50.60\pm2.23 $& $48.13\pm2.32 $& $62.15\pm2.63 $& $52.50\pm2.44 $& $56.36\pm2.73 $\\
 & $52.90\pm2.37 $ & $53.59\pm2.01 $& $52.31\pm2.91 $& $53.62\pm2.69 $& $58.70\pm2.43 $& $54.12\pm2.62 $\\
\hline

 & $54.88\pm1.53 $ & $53.58\pm1.69 $& $59.01\pm1.82 $& $61.66\pm2.13 $& $60.75\pm2.56 $& $62.47\pm2.26 $\\
MCAT & $52.33\pm2.04 $ & $56.76\pm2.31 $& $48.65\pm1.92 $& $65.33\pm2.87 $& $56.22\pm2.37 $& $60.45\pm2.74 $\\
 & $57.73\pm1.72 $ & $54.89\pm1.90 $& $50.16\pm2.11 $& $60.94\pm1.78 $& $50.27\pm2.07 $& $48.73\pm1.92 $\\
\hline
 & $55.87\pm 2.31 $ & $55.37\pm2.14 $& $43.69\pm3.52 $& $64.48\pm2.01 $& $60.89\pm2.36 $& $60.34\pm2.29 $\\
\makecell{PORP\\-OISE} & $55.40\pm2.57 $ & $50.20\pm2.43 $& $53.51\pm2.46 $& $54.21\pm2.83 $& $42.67\pm2.74 $& $45.23\pm2.72$\\
 & $50.70\pm2.34 $ & $54.95\pm2.37 $& $49.48\pm2.17 $& $63.00\pm2.05 $& $55.27\pm2.41 $& $50.89\pm2.10 $\\
\hline
\end{tabular}
\end{table}

\subsection{The Results of Baseline Models with Fewer Finetuning Data}
In Table~\ref{tabS2}, for each baseline, we include three rows to show the C-index values. The three rows are shown as "image and mRNA", "image and CNA", and "image and Methylation" in order. The baseline models are trained by fewer finetuning multimodal data.



\end{document}